\documentclass[preprint]{elsarticle}
\usepackage{lineno,hyperref,color}
\usepackage{subfigure}
\modulolinenumbers[5]
\journal{Journal of \LaTeX\ Templates}
\usepackage{amsfonts}
\usepackage{amsmath}
\usepackage{amsthm}
\usepackage{amssymb}
\usepackage{bm}
\usepackage{enumerate}
\usepackage{multirow}
\usepackage{dsfont}
\usepackage{mathrsfs}
\usepackage{booktabs}
\usepackage{graphicx,epstopdf}
\usepackage{color}
\usepackage{clrscode}

\DeclareMathAlphabet{\mathsfsl}{OT1}{cmss}{m}{sl}
\definecolor{orange}{RGB}{255,127,0}

\newtheorem{thm}{Theorem}

\newdefinition{rmk}{Remark}
\newdefinition{definition}{Definition}

\graphicspath{{./Figures/}} 

\begin{document}

\begin{frontmatter}

\title{How priors of initial hyperparameters affect Gaussian process regression models}

\author {Zexun Chen\corref{cor1}}
\ead{zc74@le.ac.uk}
\cortext[cor1]{Corresponding author. Tel: +44-(0)77-6794-0465}

\author{Bo Wang\corref{}}
\address{Department of Mathematics, University of Leicester, Leicester, LE1 7RH, UK}
%
%
\begin{abstract}
The hyperparameters in Gaussian process regression (GPR) model with a specified kernel are often estimated from the data via the maximum marginal likelihood. Due to the non-convexity of marginal likelihood with respect to the hyperparameters, the optimization may not converge to the global maxima. A common approach to tackle this issue is to use multiple starting points randomly selected from a specific prior distribution. As a result the choice of prior distribution may play a vital role in the predictability of this approach. However, there exists little research in the literature to study the impact of the prior distributions on the hyperparameter estimation and the performance of GPR. In this paper, we provide the first empirical study on this problem using simulated and real data experiments. We consider different types of priors for the initial values of hyperparameters for some commonly used kernels and investigate the influence of the priors on the predictability of GPR models. The results reveal that, once a kernel is chosen, different priors for the initial hyperparameters have no significant impact on the performance of GPR prediction, despite that the estimates of the hyperparameters are very different to the true values in some cases.
\end{abstract}


\end{frontmatter}

\section{Introduction}
Over the last few decades, Gaussian Processes Regression (GPR) has been proven to be a powerful and effective method for non-linear regression problems due to many desirable properties, such as ease of obtaining and expressing uncertainty in predictions, the ability to capture a wide variety of behaviour through a simple parameterization, and a natural Bayesian interpretation \cite{cite:RCE}.
Neal \cite{cite:neal} revealed that many Bayesian regression models based on neural network converge to Gaussian Processes (GPs) in the limit of an infinite network. Therefore, GPs have been suggested as a replacement for supervised neural networks in non-linear regression \cite{cite:MacKay,cite:WCKI} and classification \cite{cite:MacKay}. In recent years, various empirical studies have shown that GPR can make better performance for prediction in many areas \cite{he2011single,ko2009gp,van2002bayesian,yu2012online} compared to some other models such as Support Vector Machine (SVM) \cite{shamshirband2015comparative,jahangirzadeh2014cooperative,shamshirband2015soft}, and a number of further developments of Gaussian process methods have been proposed, including deep Gaussian process \cite{damianou2013deep} and recurrent Gaussian processes \cite{mattos2015recurrent}.

However, GPR as a kernel-based nonparametric method, relies on appropriate selection of kernel \cite{wilson2013gaussian} and the hyperparameters involved. Kernels contain our presumptions about the function we wish to learn and define the closeness and similarity between data points \cite{cite:Rbook}. As a result, the choice of kernel has a profound impact on the performance of a GPR model, just as activation function, learning rate can affect the result of a neural network \cite{wilson2013gaussian}.

Once a kernel is selected for a kernel-based method, the unknown hyperparameters involved in the kernel need to be estimated from the training data. Although Monte Carlo methods can perform GPR without the need of estimating hyperparameters \cite{cite:WCKI,cite:monte,cite:BBS,mackay1998introduction}, the common approach is
to estimate the hyperparameters by means of maximum marginal likelihood \cite{cite:MacKay} due to the high computational cost of Monte Carlo methods. Unfortunately marginal likelihood functions are not usually convex with respect to the hyperparameters, which means local optima may exist \cite{cite:autokernel} and the optimized hyperparameters, which depend on the initial values, may not be the global optima \cite{cite:WCKI,cite:BBS,mackay1998introduction,wilson2014covariance}.
A common approach to tackle this issue is to use multiple starting points randomly selected from a specific prior distribution and after convergence choose the optimised values with the largest marginal likelihood as the estimates.
Most practitioners using GPR as a modelling tool tend to choose a simple prior distribution based on their expert opinions and experiences, such as the Uniform distribution in the range of (0, 1) \cite{cite:WCKI,cite:BBS,wilson2014covariance}.
However, it is of importance and of interest to investigate whether the predictability of GPR models would be jeopardised if the prior distribution were not properly chosen and how the choice of prior distribution may affect the performance of GPR models \cite{cite:autokernel,wilson2014covariance}. If the performance of GPR is sensitive to the choice of prior distribution, the prior needs to be chosen carefully when using GPR models; otherwise, a simple prior may be sufficient.
The study of this kind could provide useful guidances to researchers and practitioners using GP as a modelling tool.

This paper provides the first empirical study on this problem using simulated and real data experiments. We consider different types of priors, including vague and data-dominated, for the initial values of hyperparameters for some commonly used kernels and investigate the influence of the priors on the predictability of GPR models.
The paper is organized as follows. Section 2 is a brief introduction to GPR. In Section 3, we discuss the key problem of the sensitivity of initial hyperparameters. Section 4 describes some different prior distributions for initial values, including both non-informative and data-dominated priors. Numerical experiments for different samples, including simulated data and real data, over different kernels are demonstrated and discussed in Section 5. Section 6 concludes the paper.

\section{Background}
\subsection{Gaussian processes regression model}
A Gaussian process is a collection of random variables, any finite number of which have (consistent) Gaussian distribution.

Mathematically, for any set $S$ \footnote{Although $S$ can be any set, it usually is $\mathds{R}$ or $\mathds{R}^n$. In this paper, we consider $S = \mathds{R}$ only.}, a Gaussian process (GP) on $S$ is a set of random variables ($f_x, x\in S$) such that, for any $n \in\mathds{N}$ and $x_1,\ldots, x_n \in S, (f_{x_1},\ldots,f_{x_n})$ is (multivariate) Gaussian.

As a Gaussian distribution is specified by a mean vector and a covariance matrix, a GP is also fully determined by a mean function and a covariance function. In other words, we have:
\begin{thm}[Gaussian Processes]
For any set $S$, any mean function $\mu : S\mapsto \mathds{R}$ and any covariance function (also called kernel) $k: S\times S \mapsto \mathds{R}$, there exists a GP $f(x)$ on $S$, $s.t. $ $\mathbb{E}[f(x)] = \mu(x)$, $Cov(f(x_s), f(x_t)) = k(x_s,x_t), \forall x, x_s,x_t \in S$. It denotes $f \sim \mathcal{GP}(\mu, k)$.
\end{thm}

For a regression problem $y=f(x)+\varepsilon$, by Gaussian process method
the unknown function $f$ is assumed to follow a $\mathcal{GP}(\mu,k)$.
Given $n$ pairs of observations $(x_1, y_1), \ldots, (x_n, y_n)$, we have
 $\bm{y} = f(X) + {\boldsymbol\varepsilon} $, where $\bm{y} = [y_1,y_2,\ldots,y_n]^{\mathrm{T}}$ are the outputs, $X = [x_1,x_2,\ldots,x_n]^{\mathrm{T}}$ are the inputs, and
${\boldsymbol\varepsilon= [\varepsilon_1,\varepsilon_2,\ldots,\varepsilon_n]^{\mathrm{T}}}$ are
independent identically distributed Gaussian noise with mean 0 and variance $\sigma_n^2$
\cite{cite:Rbook}. It yields that the collection of functions $[f(x_1),\ldots,f(x_n)]$ follow a multivariate Gaussian distribution

$$
[f(x_1),f(x_2),\ldots,f(x_n)]^{\mathrm{T}} \sim \mathcal{N}(\bm{\mu},K),
$$
where $\bm{\mu} = [\mu(x_1),\ldots,\mu(x_n)]^\mathrm{T}$ is the mean vector and $K$ is the $n \times n$ covariance matrix of which the $(i,j)$-th element $K_{ij} = k(x_i,x_j)$.

To predict the function values $\bm{f}_* = [f_{*1},\ldots,f_{*m}]^\mathrm{T}$ at the test locations $X_* = [x_{n+1},\ldots,x_{n+m}]^\mathrm{T}$, the joint distribution of training observations $\bm{y}$ and predictive targets $\bm{f}_*$ are given by
\begin{equation}\label{joint}
  \begin{bmatrix}
  \bm{y}  \\
  \bm{f}_*
  \end{bmatrix} \sim
  \mathcal{N}    \left(
  \begin{bmatrix}
  \mu(X) \\
  \mu(X_*)
  \end{bmatrix},
  \begin{bmatrix}
  K(X,X) + \sigma^2_n I   \quad K(X_*,X)^{\mathrm{T}} \\
  K(X_*,X)                \quad \qquad K(X_*,X_*)
  \end{bmatrix}   \right),
\end{equation}
where $\mu(X) = \bm{\mu}$, $\mu(X_*) = [\mu(x_{n+1}),\ldots,\mu(x_{n+m})]^\mathrm{T}$, $K(X,X)= K$,
$K(X_*,X)$ is an $m \times n$ matrix of which the $(i,j)$-th element
$[K(X_*,X)]_{ij} = k(x_{n+i},x_j)$, and $K(X_*,X_*)$ is an $m \times m$ matrix with
 the $(i,j)$-th element $[K(X_*,X_*)]_{ij} = k(x_{n+i},x_{n+j})$. Thus the predictive distribution is
 \begin{equation}\label{predictive}
   p(\bm{f}_*|X,\bm{y},X_*) =    \mathcal{N}(\hat{\mu},\hat{\Sigma}),
 \end{equation}
\begin{align}
  \hat{\mu}     &= K(X_*,X)^{\mathrm{T}}(K(X,X) + \sigma^2_n I)^{-1}(\bm{y}-\mu(X)),\label{predictive1} \\
  \hat{\Sigma}  &= K(X_*,X_*)  - K(X_*,X)^{\mathrm{T}}(K(X,X) + \sigma^2_n I)^{-1}K(X_*,X) \label{predictive2}.
  \end{align}
In GPR method the mean function $\mu(x)$ is often assumed to be 0, then the predictive mean and variance can be given as
  \begin{align}
  \hat{\mu}     &= K(X_*,X)^{\mathrm{T}}(K(X,X) + \sigma^2_n I)^{-1}\bm{y},\label{0mean} \\
  \hat{\Sigma}  &= K(X_*,X_*)  - K(X_*,X)^{\mathrm{T}}(K(X,X) + \sigma^2_n I)^{-1}K(X_*,X) \label{0varaince}.
  \end{align}

\subsection{Kernels}
From the view of Eq.\eqref{0mean} and Eq.\eqref{0varaince}, the kernel $k(\cdot,\cdot)$ plays a crucial role in the predictive mean and variance. As discussed in \cite{cite:Rbook}, kernels contain our presumptions about the function we wish to learn and define the closeness and similarity between data points. As a result, the choice of kernel has a profound impact on the performance of a GPR model, just as activation function, learning rate can affect the result of a neural network \cite{wilson2013gaussian}. Some commonly used kernels are listed as follows.

\subsubsection{Squared exponential}
The most widely-used kernel in GPR is Squared Exponential (SE), which is defined as
$$
k_{SE}(x,x') = s_f^2 \exp (- \frac{(x-x')^2}{2\ell^2}),
$$
where $s_f$ is the signal variance and can also be considered as an output-scale amplitude \cite{cite:twins} and the parameter $\ell$ is the input (length or time) scale \cite{cite:twins}.

\subsubsection{Periodic}
Periodic kernel (PER) is used to model functions which exhibit a periodic pattern. It is given by
$$
k_{PER}(x,x') = s_f^2 \exp(- \frac{2\sin^2(\pi \frac{(x - x')}{p})}{\ell^2}),
$$
where $p$ is the period of the function and the parameters $s_f$ and $\ell$
have the same meaning as in SE.

\subsubsection{Local periodic}
As seen in \cite{cite:autokernel}, positive semi-definite kernels are closed under addition and multiplication. Local Periodic (LP) is such a composite kernel which is obtained
by multiplying SE and PER \cite{cite:autokernel}. That is,
$$
k_{LP}(x,x') = k_{SE}(x,x') \times k_{PER}(x,x').
$$
It is a well-known kernel to capture locally periodic structure of data hence can be applied to many kernel-based models.

\subsubsection{Spectral mixture}
The Spectral Mixture (SM) kernel was introduced by Wilson \cite{wilson2014covariance} and is defined
as a scaled mixture of Q Gaussians:
$$
k_{SM}(x,x') = \sum^Q_{q=1}w_q \exp(-2\pi^2(x - x')^2 \nu_q )\cos(2\pi(x -x')\mu_q),
$$
where $w_q$'s are the weights, the inverse means $1/\mu_q$ represent component period and each inverse standard deviation $1/\sqrt{\nu_q}$ represents a length scale \cite{wilson2014covariance}.

\subsection{Estimation of hyperparameters}

In GPR models, the hyperparameters involved in the kernel need to be estimated from the training data. Although Monte Carlo methods can perform GPR without the need of estimating hyperparameters \cite{cite:WCKI,cite:monte,cite:BBS,mackay1998introduction}, the common approach is
to estimate them by means of maximum marginal likelihood due to the high computational cost of Monte Carlo methods.

Following the GP assumption, the distribution of the training outputs is given as
\begin{equation}\label{parameter}
  p(\bm{y}|X,\bm{\theta}) = \mathcal{N}(\bm{0},\Sigma_{\theta}),
\end{equation}
where $\Sigma_{\theta} = K+\sigma^2_n I$ and $ \bm{\theta}$ is the collection of the unknown hyperparameters.
Therefore, the negative log marginal likelihood (nlml) is
\begin{equation}\label{nlml}
  \mathcal{L}(\bm{\theta})
  = -\log p(\bm{y}|X,\bm{\theta})
  = \frac{1}{2}\bm{y}^{\mathrm{T}}  \Sigma^{-1}_{\theta}\bm{y}  +\frac{1}{2} \log \det \Sigma_{\theta}  + \frac{n}{2}\log 2\pi,
\end{equation}
and the partial derivatives of nlml with respect to the hyperparameters are given by
\begin{equation}\label{deerivative}
  \frac{\partial}{\partial \theta_i}\mathcal{L}(\bm{\theta}) = \frac{1}{2} \mathrm{tr}(\Sigma_{\theta}^{-1} \frac{\partial \Sigma_{\theta}}{\partial \theta_i}) - \frac{1}{2} \bm{y}^{\mathrm{T}}  \Sigma_{\theta}^{-1}\frac{\partial \Sigma_{\theta}}{\partial \theta_i}\Sigma_{\theta}^{-1}\bm{y}.
\end{equation}


\section{Sensitivity of prior distributions for initial hyperparameters}
For many kernels the likelihood function \eqref{nlml} is not convex with respect to the  hyperparameters, therefore the optimisation algorithm may converge to a local optimum whereas the global one may provide better results \cite{cite:autokernel}. As a result the optimised hyperparameters achieved by maximum likelihood estimation and the performance of GPR may depend on the initial values of the optimisation algorithm \cite{cite:WCKI,cite:BBS,mackay1998introduction,wilson2014covariance}.

A common strategy adopted by most GPR practitioners is a heuristic method. That is, the optimisation
is repeated using several initial values generated randomly from a simple prior distribution,
which is often selected based on their expert opinions and experiences.
The final estimates of the hyperparameters are the ones with the largest likelihood values after
convergence \cite{cite:WCKI,cite:BBS,wilson2014covariance}. It is therefore interesting to
know how prior distributions affect the performance of GPR since the above strategy can not guarantee a global maximum of the likelihood function is found, or the sensitivity of prior distributions to the performance of GPR, which, to the best of our knowledge, has not been studied in the literature. In this paper, we provide the first empirical study of the impact of the prior distributions on the hyperparameter estimation and the performance of GPR, for some commonly used kernels in GPR modelling. The procedure for hyperparameter estimation is described below.

\begin{codebox}
\Procname{$\proc{Hyperparameter estimation}$}
Given a prior distribution $p_0(\bm{\theta})$ and the number of repetitions $M$ \\

\li Randomly choose an initial hyperparameter $\bm{\theta}_0$ from $p_0(\bm{\theta})$
\li Numerically minimise $\mathcal{L}(\bm{\theta})$ in \eqref{nlml} using $\bm{\theta}_0$ as the starting value and obtain \\
an estimate of the hyperparameter
\li Repeat Steps 1 and 2 for $M$ times and select the estimate with the smallest \\
negative log marginal likelihood as the optimal estimate
\End
\end{codebox}

Prediction is then made based on the optimal estimate of the hyperparameter
and the prediction accuracy for different priors $p_0(\bm{\theta})$ are compared.

\section{Prior distributions of initial hyperparameters}
The prior distributions considered include non-informative \cite{wilson2013gaussian} and data-dominated \cite{wilson2014covariance}, which are briefly introduced as follows.

\subsection{Vague priors}
In the cases when there is little information about the data, vague prior distributions are often selected with the intention that they should have slight or no influence on the inferences \cite{lambert2005vague,gelman2006prior}. Many justifications and interpretations of non-informative priors have been proposed over the years, such as maximum entropy \cite{rosenkrantz2012jaynes}. However, with small amount of data, the use of non-informative prior may be problematic and a vague prior distribution may lead to significant influence on any inference made because the results are easily sensitive to the selection of prior distributions
 \cite{lambert2005vague}.

Let $\theta_i$ be a generic notation for a hyperparameter in a given kernel. Below list the weakly-informative prior distributions which will be discussed in our study.

\textbf{Prior 1}
$$
\theta_i \sim \mathrm{Uniform}(0,1).
$$
This is probably the most common prior distribution. Actually, it is not strictly a `vague' prior since the range of the distribution is restricted. However, this prior is very widely-used for the estimation of the unknown parameters in the GPR models.

\textbf{Prior 2}
$$
\log(\theta_i) \sim \mathrm{Uniform}(-1,1).
$$
This prior distribution is uniform on the log hyperparameters in $(-1,1)$, so the range of the hyperparameters is $(1/\mathrm{e}, \mathrm{e})$.

\textbf{Prior 3}
$$
\log(\theta_i) \sim \mathrm{Uniform}(-10,10).
$$
This prior is similar to Prior 2 but has much larger range. So the range of the hyperparameters
 is approximately $(0, \mathrm{e}^{10})$.

\textbf{Prior 4}
$$
\theta_i \sim \mathcal{N}(0,1).
$$
The standard normal prior is also a popular and simple choice. It is not strictly a `vague' prior either, and cannot be used for positive parameters.

\textbf{Prior 5}
$$
\frac{\pi}{\theta_i} \sim \mathrm{Uniform}(0,1).
$$
This prior is specified for the period parameter for kernels that contain periodic part.
The range of the parameter is $(\pi, +\infty)$.

\textbf{Prior 6}
$$
\log(\frac{\pi}{\theta_i}) \sim \mathrm{Uniform}(-5,5).
$$
This prior is also specified for the period parameter. It is similar to Prior 5 but with a
range $(\pi\mathrm{e}^{-5}, \pi\mathrm{e}^5)$.

\subsection{Data-dominated priors}
Data-dominated priors are incorporated with some information inferred from training data, such as the possible range of the initial hyperparameters. The following data-dominated priors will be
used in this study.

\textbf{Prior 7}
$$
\theta_i \sim \mathrm{Uniform}(0,\mathrm{Nyq}).
$$
This prior is also specified for the period parameter and is based on Nyquist frequency \cite{bammes2012direct}, where Nyq equals half the sampling rate of the data, or half the largest interval between input points if the data are not regularly sampled \cite{wilson2014covariance}. Nyquist frequency can be used to find the
approximate period of data in signal processing and spectral analysis.
For example, Wilson \cite{wilson2014covariance} used this prior to initialise SM kernel.

\textbf{Prior 8}
$$
\frac{1}{\theta_i} \sim \mathcal{TN}(\mathrm{MaxI}),
$$
where $\mathcal{TN}(\mathrm{MaxI})$ is the truncated normal distribution with mean proportional to the maximal range of the inputs (MaxI) \cite{wilson2014covariance}. It is
an improved version of Prior 4 and is used by Wilson \cite{wilson2014covariance} for the length scale in SM kernel.

\textbf{Prior 9}
$$
\frac{\pi}{\theta_i} \sim \mathrm{Uniform}(\frac{\pi}{\mathrm{MaxI}},\mathrm{\pi Nyq}).
$$
This prior is also specified for the period parameters and has the range from $1/ \mathrm{Nyq}$ to $\mathrm{MaxI}$. 

\section{Experiments}
\subsection{Experiments using samples from Gaussian processes}
In this section, we study how the priors of initial hyperparameters affect the estimates of the hyperparameters and the performance of GPR models using data generated from specified Gaussian processes. Since the true models are known, the accuracy of the estimates can be compared.


Letting $x_i=i$ for $i = 1,2,\ldots,400$, we generate samples $\{y_i\}$ from GPs with zero mean and SE and PER kernels, respectively. These two kernels are used as demonstration because SE is the most widely-used kernel in GPR while PER is the simplest kernel which may suffer from the problem of local optima in optimisation procedure because integer multiples of the true period, such as harmonics, are often local optima \cite{cite:autokernel}.

To evaluate the influences of the prior distributions on the hyperparameter estimation, ten values randomly generated from each prior distribution discussed in Section 4 (where applicable) are used as the starting values for the maximum likelihood procedure, implemented by Conjugate Gradient
algorithm. Among the ten estimates after the procedure is converged the one with the largest maximum likelihood is chosen as the optimal estimate, denoted by $\bm{\theta}_{final}$, and is compared with
 $\bm{\theta}_{act}$.

To study the impact of the priors on the predictability of GPR, we consider two types of prediction:
interpolation and extrapolation. Denote the whole data set
by $\Omega = \{(i,y_i); i = 1,2,\ldots,400 \}$. For interpolation, the test set is given by
 $\mathcal{D}_{I2} =  \{(i,y_i); i = 5j+1, j=0,1,\ldots,79 \}$ and the training set is $\mathcal{D}_{I1} = \Omega - \mathcal{D}_{I2} $. For extrapolation, the training set is $\mathcal{D}_{E1} =  \{(i,y_i); i = 1,2,\ldots,320 \}$ and the test set is $\mathcal{D}_{E2} = \Omega - \mathcal{D}_{E1}$.

The predicted values are then compared with the actual values. There are several ways to evaluate the accuracy of the predictions. The simplest one is the root mean squared error (RMSE), which is defined as
$$
 \mbox{RMSE}     = \sqrt{\frac{1}{m}\sum^m_{i=1}(\hat{y}_i - y_{i})^2},
$$
where $\{\hat y_i\}$ and $\{y_i\},i=1,2,\ldots,m$, are the predicted mean values and the actual test values respectively. However, the RMSE can be affected seriously by the overall scale of the output values, so we utilize the standardized root mean squared error (SRMSE) which is normalized by the standard deviation of $\{y_i\}$, i.e.
$$
\mbox{SRMSE} = \frac{\mbox{RMSE}}{\sigma_y},
$$
where $\sigma_y$ is the standard deviation of $\{y_i\},i=1,2,\ldots,m$. This implies any model which can provide the prediction close to the sample mean of the test targets to have a SRMSE of approximately 1 \cite{cite:Rbook}. In other words, any prediction model with the SRMSE around 1 is satisfactory.

Another measure which can take account of both predictive mean and predictive variance is log loss. As the predictive distribution for each test point is Gaussian, its log loss is defined as
$$
\mbox{LL} = \frac{1}{2}\log(2\pi\hat{\sigma}^2_i) + \frac{(y_i - \hat{y}_i)^2}{2\hat{\sigma}_i^2},
\; i = 1,2,\ldots,m,
$$
where $\{\hat \sigma_i\},i =1,2,\ldots,m$ are the predictive variances. This loss can be standardized by subtracting the loss that could be obtained by the null model which predicts using a Gaussian with the sample mean and sample variance of the training outputs \cite{cite:Rbook}. And the mean standardized log loss (MSLL) is the average of the standardized log loss for $i = 1,2,\ldots,m$. Therefore, the MSLL is zero for null model, and the smaller it is the better a model is in terms of loss \cite{cite:Rbook}.

\subsubsection{Squared Exponential kernel}
As can be seen in Section 4, not all of the priors are suitable for every hyperparameter.
Therefore for SE kernel, we use \textbf{Prior} 1, \textbf{Prior} 2, \textbf{Prior} 3 for
both hyperparameters $[\ell,s_f]$. The data are generated using
 $\bm{\theta}_{act} = [\ell,s_f]= [5, 2]$.

\begin{figure}[htbp]
\centering
\includegraphics[width=.6\textwidth]{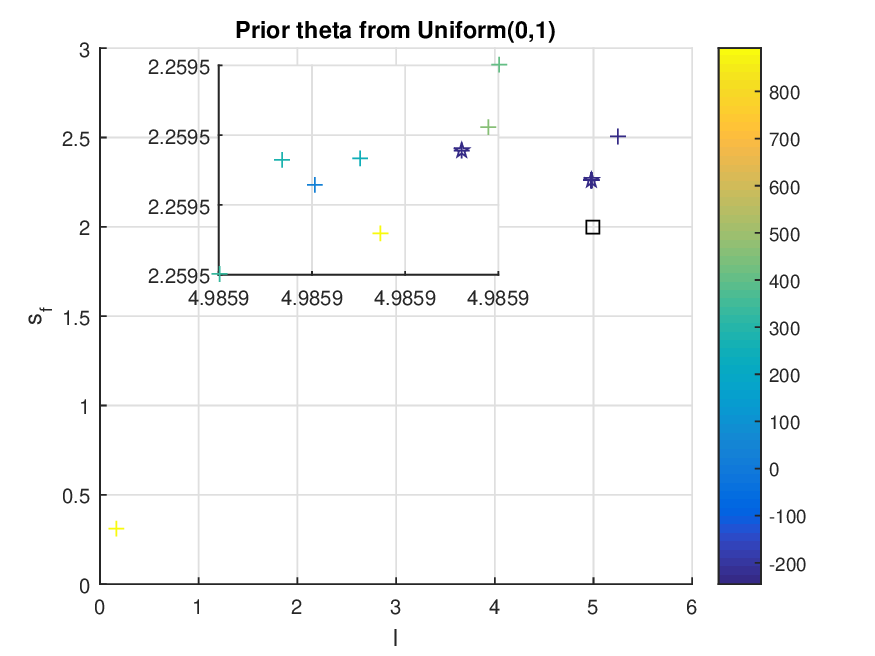}
\includegraphics[width=.6\textwidth]{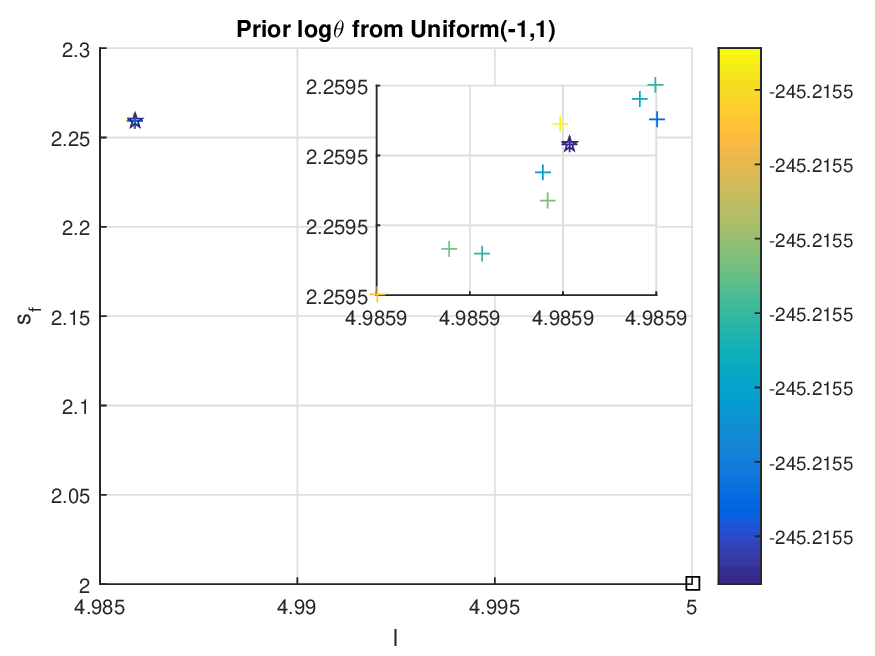}
\includegraphics[width=.6\textwidth]{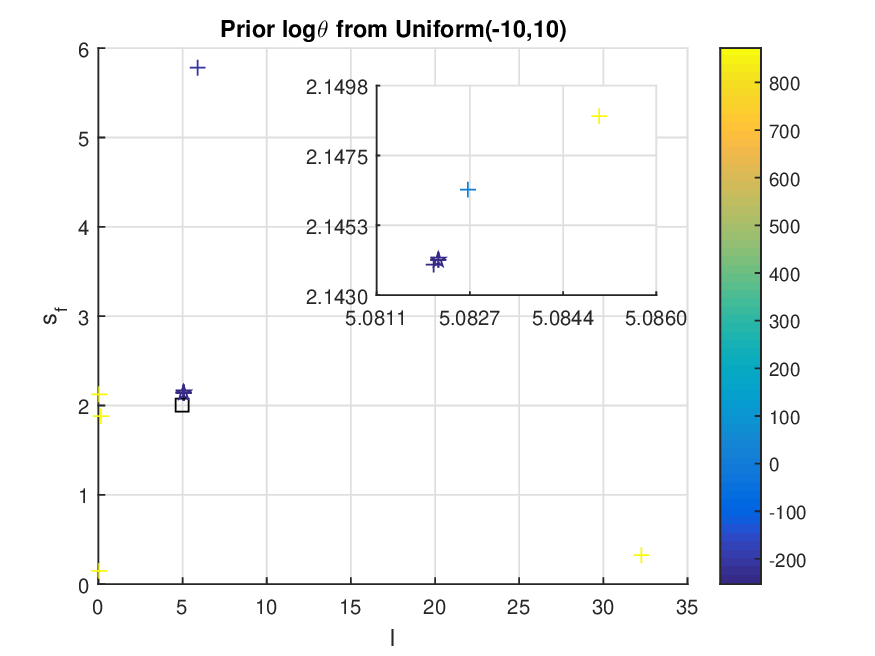}
\caption{Positions of the estimated hyperparameters for the SE kernel. Top to bottom: Priors 1, 2 and 3.}
\label{SEprior123}
\end{figure}

To compare $\bm{\theta}_{act}$ and $\bm{\theta}_{final}$, Figure \ref{SEprior123} illustrates their
visual positions, where $``\Box"$ represents $\bm{\theta}_{act}$ ,
 $``\star"$ represents $\bm{\theta}_{final}$, the $``+"$s are the intermediate values during the process of optimization and
 the color of the symbols stands for the value of the negative log marginal likelihood (nlml).

Apparently, regardless of the priors, the optimisation converges very fast and
the estimated hyperparameter $\bm{\theta}_{final}$ is always very close to $\bm{\theta}_{act}$.\\

We now test the prediction performance by the GPR with SE kernel. Only \textbf{Prior 1} is used
since the estimated hyperparameters from different priors are almost the same. The samples are
generated using two GP models with two different hyperparameters: $\bm{\theta}_{act} = [5, 2]$ and $\bm{\theta}_{act} =[15,7]$, respectively. And the above experiment is repeated 20 times and the average results are reported in Table \ref{tab:addlabel}.
It is obvious that the mean estimate
of $\bm{\theta}_{final}$ is very close to $\bm{\theta}_{act}$ with small standard errors for
both cases, and the GPR model performs well and stably for both interpolation and extrapolation predictions.
\begin{table}[htbp]
   \small
  \centering
  \caption{Results of GP predictions with SE kernel (the standard errors are given in the brackets)}
    \begin{tabular}{rrrccc}
    \toprule
          & \multicolumn{2}{c}{} & \multicolumn{3}{c}{Interpolation} \\
    \midrule
    \multicolumn{1}{c}{} & \multicolumn{2}{c}{$\bm{\theta}_{act}$} & $\bm{\theta}_{final}$  & SRMSE  & MSLL \\
    \midrule
    \multicolumn{1}{c}{\multirow{2}[4]{*}{}} & \multicolumn{1}{c}{$\ell$} & 5     & 4.97 (0.133) & \multirow{2}[4]{*}{0.03 (0.004)}
		& \multirow{2}[4]{*}{-3.48 (0.130) } \\
    \multicolumn{1}{c}{} & \multicolumn{1}{c}{$s_f$} & 2     & 2.00 (0.187) &       &  \\
    \midrule
    \multicolumn{1}{c}{\multirow{2}[4]{*}{}} & \multicolumn{1}{c}{$\ell$} & 15    & 14.95 (0.467) & \multirow{2}[4]{*}{0.01 (0.002)}
		& \multirow{2}[4]{*}{-4.74 (0.229)} \\
    \multicolumn{1}{c}{} & \multicolumn{1}{c}{$s_f$} & 7     & 6.94 (1.110) &       &  \\
    \midrule
          &       &       & \multicolumn{3}{c}{Extrapolation} \\
    \midrule
    \multicolumn{1}{c}{} & \multicolumn{2}{c}{$\bm{\theta}_{act}$} & $\bm{\theta}_{final}$  & SRMSE  & MSLL \\
    \midrule
    \multicolumn{1}{c}{\multirow{2}[4]{*}{}} & \multicolumn{1}{c}{$\ell$} & 5     & 4.98 (0.165) & \multirow{2}[4]{*}{1.02 (0.141)}
		& \multirow{2}[4]{*}{-0.14 (0.114)} \\
    \multicolumn{1}{c}{} & \multicolumn{1}{c}{$s_f$} & 2     & 1.97 (0.210) &       &  \\
    \midrule
    \multicolumn{1}{c}{\multirow{2}[4]{*}{}} & \multicolumn{1}{c}{$\ell$} & 15    & 15.01 (0.495) & \multirow{2}[4]{*}{1.19 (0.535)}
		& \multirow{2}[4]{*}{-0.56 (0.322)} \\
    \multicolumn{1}{c}{} & \multicolumn{1}{c}{$s_f$} & 7     & 7.01 (1.242) &       &  \\
    \bottomrule
    \end{tabular}%
  \label{tab:addlabel}%
\end{table}%

\subsubsection{Periodic Kernel}
Three parameters $[\ell, p, s_f]$ are involved in the PER kernel. We consider five priors (\textbf{Prior} 1, \textbf{Prior} 5, \textbf{Prior} 6, \textbf{Prior} 7 and \textbf{Prior} 9) for the $p$ term and \textbf{Prior 1} for the parameters $\ell$ and $s_f$.
In the following experiment, the data are generated using the true parameters $\bm{\theta}_{act} = [5,7,2]$.

\begin{figure}[htbp]
\centering
\subfigure[]{\includegraphics[width=.48\textwidth]{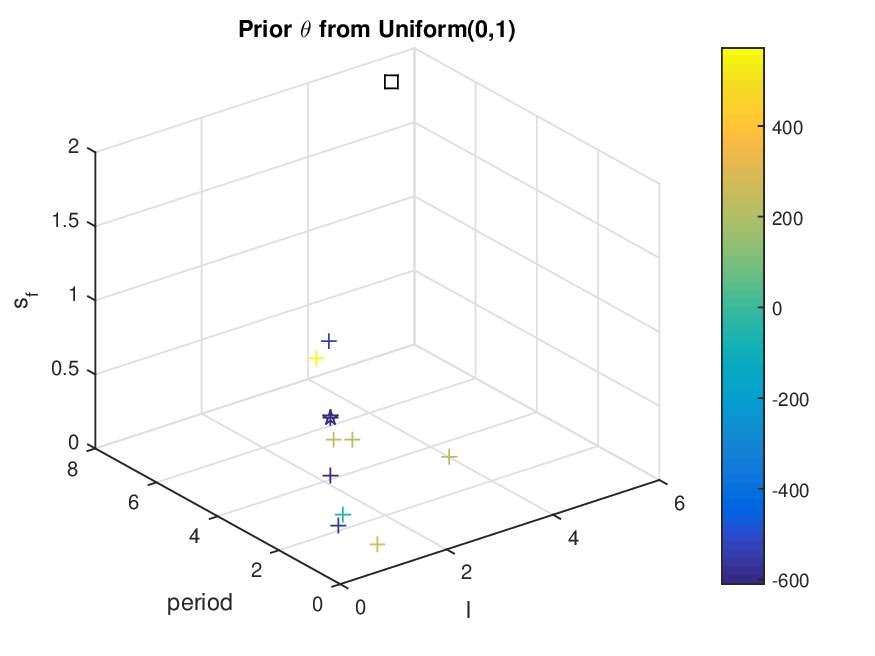}}
\subfigure[]{\includegraphics[width=.48\textwidth]{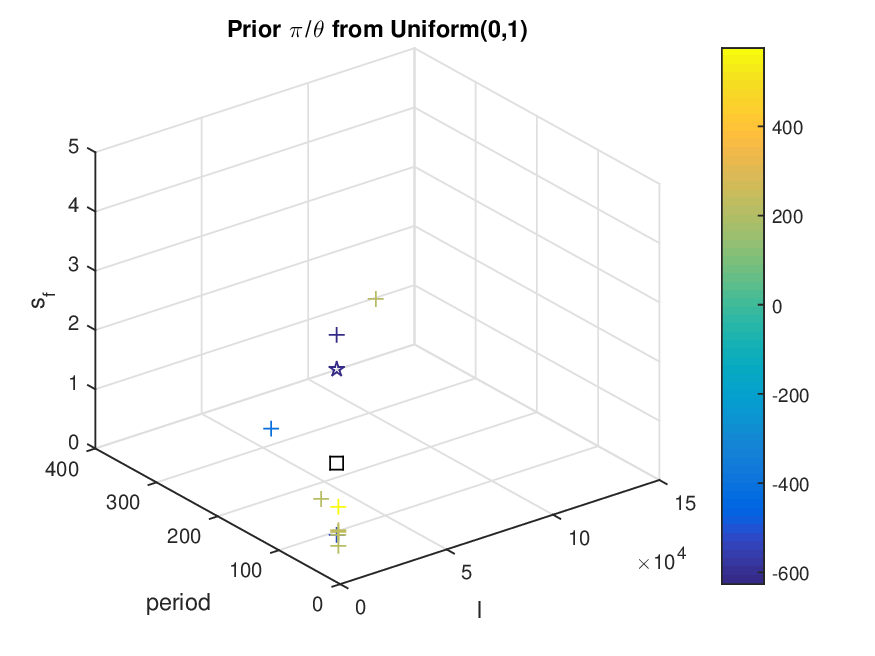}}
\subfigure[]{\includegraphics[width=.48\textwidth]{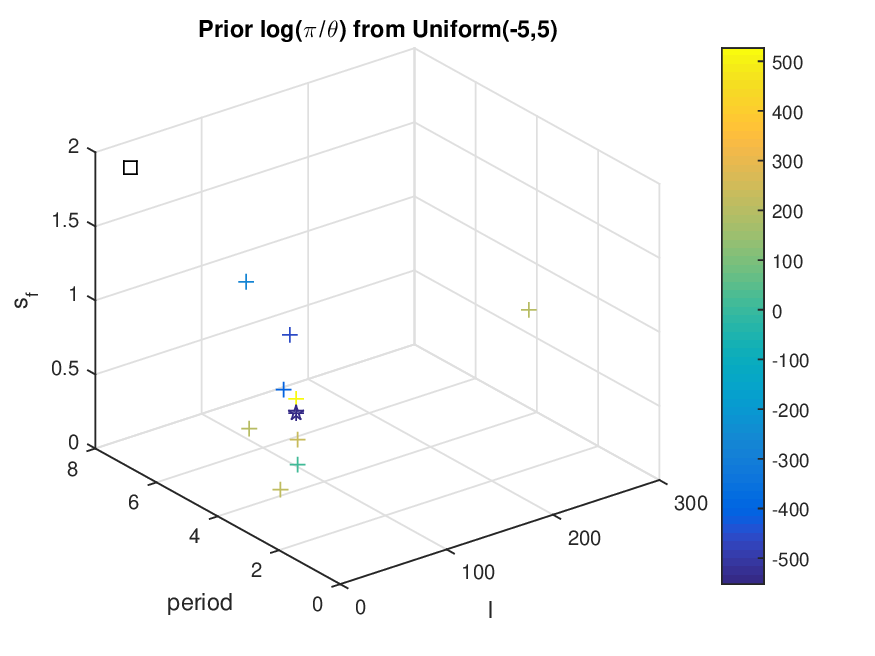}}
\subfigure[]{\includegraphics[width=.48\textwidth]{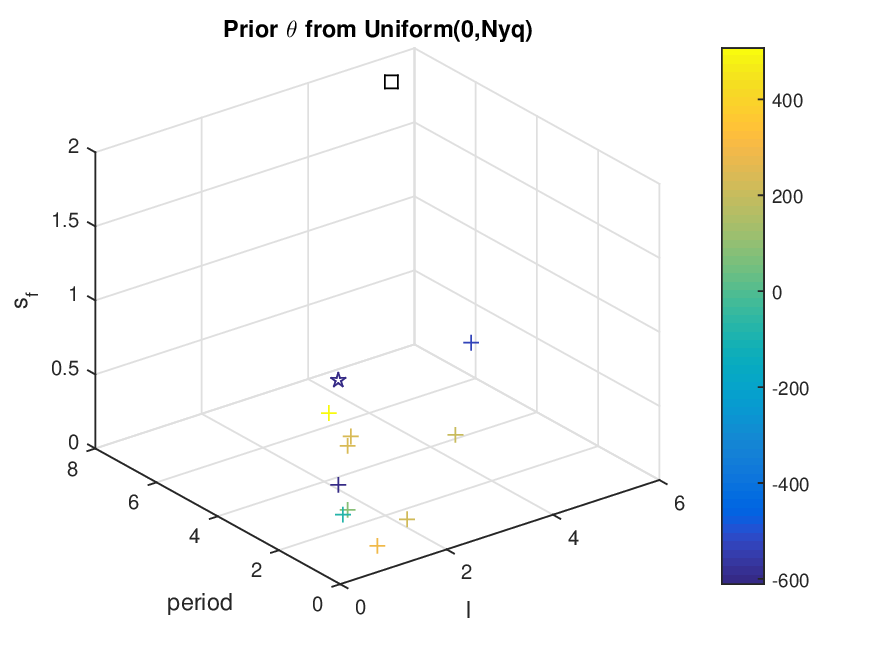}}
\subfigure[]{\includegraphics[width=.48\textwidth]{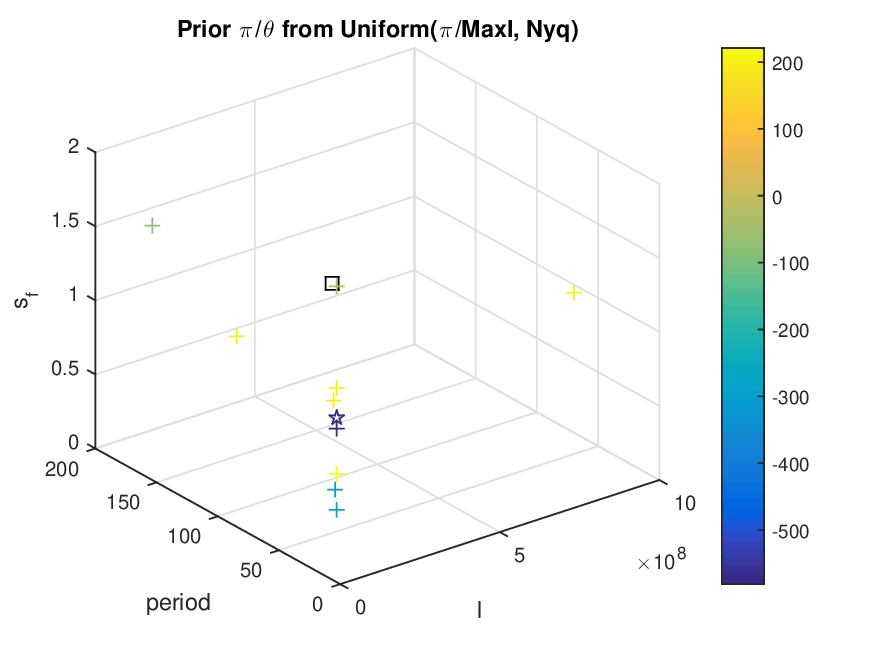}}
\caption{Positions of the estimated hyperparameters for the PER kernel. The priors for $p$ are: (a) \textbf{Prior} 1, (b) \textbf{Prior} 5, (c) \textbf{Prior} 6, (d) \textbf{Prior} 7 and (e) \textbf{Prior} 9.}
\label{PERprior15}
\end{figure}

Figure \ref{PERprior15} shows the
visual positions of $\bm{\theta}_{act}$ and $\bm{\theta}_{final}$, where the symbols have the same meanings as in Figure \ref{SEprior123}.
It can be seen that, for all the priors considered, the estimates $\bm{\theta}_{final}$ are always far away from the true value $\bm{\theta}_{act}$. Therefore, it is difficult to achieve the global maximum by the maximum marginal likelihood method for the PER kernel, and the estimates are very sensitive to prior distributions of the initial hyperparameters.

The same strategy as for the SE kernel is used to test the prediction performance by the GPR with PER kernel, and the results are reported in Table \ref{tab:addlabe2}. It can be seen that, consistent with the above findings, the averages of the estimated hyperparameters are very different than the true values, which confirms that the estimates obtained by numerical optimization of likelihood function are biased. However, both the means and standard deviations of SRMSE and MSLL are very small, which indicates that the GPR models perform very well and stably for both interpolation and extrapolation, despite the poor estimates of the hyperparameters. Therefore, although the parameter estimation for the PER kernel is sensitive to prior distributions, the GPRs still provide good results and the performance is hardly influenced by the choice of priors.

\begin{table}[htbp]
   \small
  \centering
  \caption{Results of GP predictions with PER kernel (the standard errors are given in the brackets)}
    \begin{tabular}{rcrccc}
    \toprule
          & \multicolumn{5}{c}{Interpolation} \\
    \midrule
    \multicolumn{1}{c}{Prior} & \multicolumn{2}{c}{$\bm{\theta}_{act}$} & $\bm{\theta}_{final}$ & SRMSE  & MSLL \\
     \midrule
    \multicolumn{1}{c}{\multirow{3}[6]{*}{Prior 1}} & $\ell$   & 5    & 0.25 (0.170) & \multirow{3}[6]{*}{0.35 (0.454)} & \multirow{3}[6]{*}{-1.44 (1.205)} \\
    \multicolumn{1}{c}{} & p & 7    & 1.98 (2.807)  &       &  \\
    \multicolumn{1}{c}{} & $s_f$     & 2     & 2.41 (2.453)  &       &  \\
     \midrule
    \multicolumn{1}{c}{\multirow{3}[6]{*}{Prior 5}} & $\ell$   & 5    & 4.24 (8.968)  & \multirow{3}[6]{*}{0.48 (0.815)} & \multirow{3}[6]{*}{-1.63 (1.365)} \\
    \multicolumn{1}{c}{} & p & 7    & 4.81 (1.293) &       &  \\
    \multicolumn{1}{c}{} & $s_f$     & 2     & 95.98 (205.308)  &       &  \\
     \midrule
    \multicolumn{1}{c}{\multirow{3}[6]{*}{Prior 6}} & $\ell$   & 5    & 1.19 (0.713)  & \multirow{3}[6]{*}{0.28 (0.252)} & \multirow{3}[6]{*}{-1.50 (0.705)} \\
    \multicolumn{1}{c}{} & p & 7    & 2.98 (2.288)  &       &  \\
    \multicolumn{1}{c}{} & $s_f$     & 2     & 3.58 (6.246) &       &  \\
     \midrule
    \multicolumn{1}{c}{\multirow{3}[6]{*}{Prior 7}} & $\ell$   & 5    & 1.45 (1.289)  & \multirow{3}[6]{*}{0.28 (0.253)} & \multirow{3}[6]{*}{-1.48 (0.696)} \\
    \multicolumn{1}{c}{} & p & 7    & 0.34 (0.143)  &       &  \\
    \multicolumn{1}{c}{} & $s_f$     & 2     & 1.51 (0.838) &       &  \\
     \midrule
    \multicolumn{1}{c}{\multirow{3}[6]{*}{Prior 9}} & $\ell$   & 5    & 1.67 (1.978)  & \multirow{3}[6]{*}{0.28 (0.252)} & \multirow{3}[6]{*}{-1.51 (0.712)} \\
    \multicolumn{1}{c}{} & p & 7    & 13.54 (16.614) &       &  \\
    \multicolumn{1}{c}{} & $s_f$     & 2     & 39.26 (80.484) &       &  \\
     \midrule
          & \multicolumn{5}{c}{Extrapolation} \\
     \midrule
    \multicolumn{1}{c}{Prior} & \multicolumn{2}{c}{$\bm{\theta}_{act}$} & $\bm{\theta}_{final}$ & SRMSE  & MSLL \\
     \midrule
    \multicolumn{1}{c}{\multirow{3}[6]{*}{Prior 1}} & $\ell$   & 5    & 1.23 (1.048)  & \multirow{3}[6]{*}{0.14 (0.041)} & \multirow{3}[6]{*}{-1.98 (0.287)} \\
    \multicolumn{1}{c}{} & p & 7    & 0.40 (0.254)  &       &  \\
    \multicolumn{1}{c}{} & $s_f$     & 2     & 1.43 (1.230)  &       &  \\
     \midrule
    \multicolumn{1}{c}{\multirow{3}[6]{*}{Prior 5}} & $\ell$   & 5    & 17.87 (47.950)  & \multirow{3}[6]{*}{0.24 (0.119)} & \multirow{3}[6]{*}{-1.52 (0.492)} \\
    \multicolumn{1}{c}{} & p & 7    & 7.73 (3.126) &       &  \\
    \multicolumn{1}{c}{} & $s_f$     & 2     & 56.50 (214.170)  &       &  \\
     \midrule
    \multicolumn{1}{c}{\multirow{3}[6]{*}{Prior 6}} & $\ell$   & 5    & 2.01 (2.023)  & \multirow{3}[6]{*}{0.24 (0.120)} & \multirow{3}[6]{*}{-1.51 (0.493)} \\
    \multicolumn{1}{c}{} & p & 7    & 2.20 (1.323)  &       &  \\
    \multicolumn{1}{c}{} & $s_f$     & 2     & 2.38 (3.984)  &       &  \\
     \midrule
    \multicolumn{1}{c}{\multirow{3}[6]{*}{Prior 7}} & $\ell$   & 5    & 24.18 (87.495)  & \multirow{3}[6]{*}{0.17 (0.103)} & \multirow{3}[6]{*}{-1.91 (0.538)} \\
    \multicolumn{1}{c}{} & p & 7    & 0.25 (0.117)  &       &  \\
    \multicolumn{1}{c}{} & $s_f$     & 2     & 139.47 (612.532)  &       &  \\
     \midrule
    \multicolumn{1}{c}{\multirow{3}[6]{*}{Prior 9}} & $\ell$  & 5    & 4.07 (6.696)  & \multirow{3}[6]{*}{0.18 (0.106)} & \multirow{3}[6]{*}{-1.82 (0.563)} \\
    \multicolumn{1}{c}{} & p & 7    & 5.80 (2.898) &       &  \\
    \multicolumn{1}{c}{} & $s_f$     & 2     & 4.71 (9.551) &       &  \\
    \bottomrule
    \end{tabular}%
  \label{tab:addlabe2}%
\end{table}%

\subsection{Experiments using samples from time series}
It is of interest to investigate how prior distributions of the hyperparameters influence the
predictability of GPR if the data are generated from other models.

We consider a simple time series
model ARMA(2,1) with autoregressive coefficient $[0.8,-0.45]$ and moving average coefficient $-0.5$, and
generate 400 samples $\{y_i, i = 1,2,\ldots,400\}$ with $x_i=i$ and the starting values
$y_1 = y_2 = 1$.


We consider extrapolation only as this type of prediction is more meaningful in time series modelling. We select the first 320 data points as the training data and the rest as the test data. The GPR models are applied using two composite kernels: local periodic (LP) and spectral
mixture (SM) with 4 components, both of which are known as useful kernels for data with complex pattern
\cite{wilson2014covariance}. For LP kernel, different priors are used for the $p$ parameter while \text{Prior} 1 is used for all the remaining parameters. For SM kernel, as shown in Section 2, three
parameters $[w_q,\mu_q,\nu_q]$ are involved. However, $w_q$ can be initialised as constants proportional to the standard deviation of the data \cite{wilson2014covariance}. Therefore we only focus on the remaining hyperparameters $\mu_q$ and $\nu_q$. We denote
$$
[\mu_q,\sqrt{\nu_q}] \sim
\begin{cases}
\mbox{PS51}, & \mbox{if } \mu_q \mbox{ is from \textbf{Prior }5},  \sqrt{\nu_q} \mbox{ is from \textbf{Prior }1} \\
\mbox{PS61}, & \mbox{if } \mu_q \mbox{ is from \textbf{Prior }6},  \sqrt{\nu_q} \mbox{ is from \textbf{Prior }1} \\
\mbox{PS71}, & \mbox{if } \mu_q \mbox{ is from \textbf{Prior }7},  \sqrt{\nu_q} \mbox{ is from \textbf{Prior }1} \\
\mbox{PS91}, & \mbox{if } \mu_q \mbox{ is from \textbf{Prior }9},  \sqrt{\nu_q} \mbox{ is from \textbf{Prior }1} \\
\mbox{PS58}, & \mbox{if } \mu_q \mbox{ is from \textbf{Prior }5},  \sqrt{\nu_q} \mbox{ is from \textbf{Prior }8} \\
\mbox{PS68}, & \mbox{if } \mu_q \mbox{ is from \textbf{Prior }6},  \sqrt{\nu_q} \mbox{ is from \textbf{Prior }8} \\
\mbox{PS78}, & \mbox{if } \mu_q \mbox{ is from \textbf{Prior }7},  \sqrt{\nu_q} \mbox{ is from \textbf{Prior }8} \\
\mbox{PS98}, & \mbox{if } \mu_q \mbox{ is from \textbf{Prior }9},  \sqrt{\nu_q} \mbox{ is from \textbf{Prior }8} \\
\end{cases}
$$
where PS78 is the priors used by Wilson \cite{wilson2014covariance}.

For comparison of the performance, the prediction is also performed using the true model ARMA(2,1)
with the true parameters. The experiment is repeated 20 times and the averages and the
standard deviations are reported in Tables \ref{tab:addlabel5} and \ref{tab:addlabel6}, respectively.
\begin{table}[htbp]
  \small
  \centering
  \caption{Results of GP predictions with LP kernel for ARMA data (the standard errors are given in the brackets)}
    \begin{tabular}{ccc|cc}
    \toprule
          & \multicolumn{2}{c}{GPR with LP kernel} & \multicolumn{2}{c}{ARMA(2,1)} \\
    \midrule
    Priors  & SRMSE  & MSLL  & SRMSE  & MSLL \\
    \midrule
    Prior 1 & 1.006 (0.0221) & -0.001 (0.0151) & \multirow{5}{*}{1.006 (0.0143)} & \multirow{5}{*}{-0.002(0.0077)}  \\
    Prior 5 & 1.007 (0.0223) & -0.001 (0.0154) &              &  \\
    Prior 6 & 1.006 (0.0219) & -0.001 (0.0150) &              &  \\
    Prior 7 & 1.006 (0.0219) & -0.001 (0.0150) &              &  \\
    Prior 9 & 1.005 (0.0225) & -0.002 (0.0149) &              &  \\
    \bottomrule
    \end{tabular}%
  \label{tab:addlabel5}%
\end{table}%

\begin{table}[htbp]
  \small
  \centering
  \caption{Results of GP predictions with SM kernel for ARMA data (the standard errors are given in the brackets)}
    \begin{tabular}{ccc|cc}
    \toprule
        & \multicolumn{2}{c}{GPR with SM kernel} & \multicolumn{2}{c}{ARMA(2,1)} \\
    \midrule
    \multicolumn{1}{c}{Priors} & SRMSE  & MSLL  & SRMSE  & MSLL \\
    \midrule
    PS51  & 1.008  (0.0251)  & 0.001  (0.0207)  & \multirow{8}[0]{*}{1.006 (0.0143)} & \multirow{8}[0]{*}{-0.002 (0.0077)} \\
    PS61  & 1.007  (0.0236)  & -0.001  (0.0187)  &         &  \\
    PS71  & 1.009  (0.0255)  & 0.001   (0.0210)  &         &  \\
    PS91  & 1.006  (0.0252)  & -0.002  (0.0188)  &         &  \\
    PS58  & 1.036  (0.0498)  & 0.038  (0.0415)  &          &  \\
    PS68  & 1.043  (0.0509)  & 0.036  (0.0450)  &          &  \\
    PS78  & 1.019  (0.0350)  & 0.012  (0.0351)  &          &  \\
    PS98  & 1.032  (0.0490)  & 0.028  (0.0441)  &          &  \\

    \bottomrule
    \end{tabular}%
  \label{tab:addlabel6}%
\end{table}

The results show that for both LP and SM kernels, the performance of the GPR models has no significant differences using different prior distributions, and is comparable to that by the true model. In other words, the performance of GPR models is not sensitive to the choice of prior distributions and is as good as the true model as far as this experiment concerns.

\subsection{Modelling the response surface of a catalytic oxidation process}
In this section we investigate the influence of the prior distributions of the hyperparameters
on the
predictability of GPR using a real data example.

Alcohol oxidation into the corresponding aldehydes or ketones, in particular benzyl alcohol to benzaldehyde, is one of the most significant functional group transformations in organic synthesis \cite{wang2015gaussian}. The selected catalyst, K–-Mn/C, was prepared by co-impregnating aqueous solutions of potassium and manganese nitrates onto commercially available activated carbon. The catalytic oxidation process was conducted in a bath-type lab-scale reactor. More experimental details can be found in \cite{tang2009statistical}. Our experiments are conducted to study the impact of five process factors (reaction temperature, partial pressure of oxygen, concentration of benzyl alcohol in terms of mmol diluted within 10 ml of toluene, percentage of Mn, and K:Mn ratio) on the turn over frequency (TOF) using GPR models. In accordance with the experiments in the time series example, we select Prior 1, Prior 5, Prior 6, Prior 7 and Prior 9 for the GPR model with LP kernel and PS51, PS61, PS71, PS91, PS58, PS68, PS78 and PS98 for the GPR model with SM kernel with 4 components.
Prior to the experiments, all the data are normalised by
$$
\tilde{y}_i = \frac{y_i-\mu}{\sigma},
$$
where $\mu$ and $\sigma$ are the sample mean and standard deviation of the data $\{y_i\}_{i=1}^n$ respectively.

For model training we randomly select N out of 38 data points, and the trained model is then used to make predictions for TOF on the remaining data points. Table \ref{tab:averageRMSE} presents the average RMSEs based on 20 replications using the above scheme for N = 10, 20 and 30 for
different priors. It is obvious that with the increase of the number of training points, the prediction accuracy by GPR with both kernels increases as well.
\begin{table}[htbp]%
 \caption{Average RMSEs of 20 replications for the prediction of TOF}
 \centering
  \subtable[Kernel LP]{
    \begin{tabular}{crrr}
    \toprule
    Prior & N=10  & N=20  & N=30 \\
    \midrule
    1     & 0.841 & 0.666 & 0.599 \\
    5     & 0.762 & 0.523 & 0.484 \\
    6     & 0.788 & 0.597 & 0.524 \\
    7     & 0.828 & 0.655 & 0.595 \\
    9     & 0.814 & 0.609 & 0.533 \\
    \bottomrule
    \end{tabular}%
  \label{tab:RMSElpTOF}%
 }
 \quad
 \subtable[Kernel SM]{
    \begin{tabular}{crrr}
    \toprule
    Prior & N=10  & N=20  & N=30 \\
    \midrule
    PS51  & 1.023 & 0.795 & 0.632 \\
    PS61  & 1.009 & 0.960 & 0.723 \\
    PS71  & 1.083 & 0.887 & 0.578 \\
    PS91  & 1.100 & 1.058 & 0.630 \\
    PS58  & 0.922 & 0.921 & 0.641 \\
    PS68  & 0.978 & 0.849 & 0.747 \\
    PS78  & 1.022 & 0.883 & 0.615 \\
    PS98  & 1.060 & 0.897 & 0.691 \\
    \bottomrule
    \end{tabular}%
  \label{tab:RMSEsmTOF}%
 }
  \label{tab:averageRMSE}
\end{table}

\begin{figure}[htbp]
\centering
\subfigure[Kernel LP, N = 10]{\includegraphics[width=0.49\textwidth]{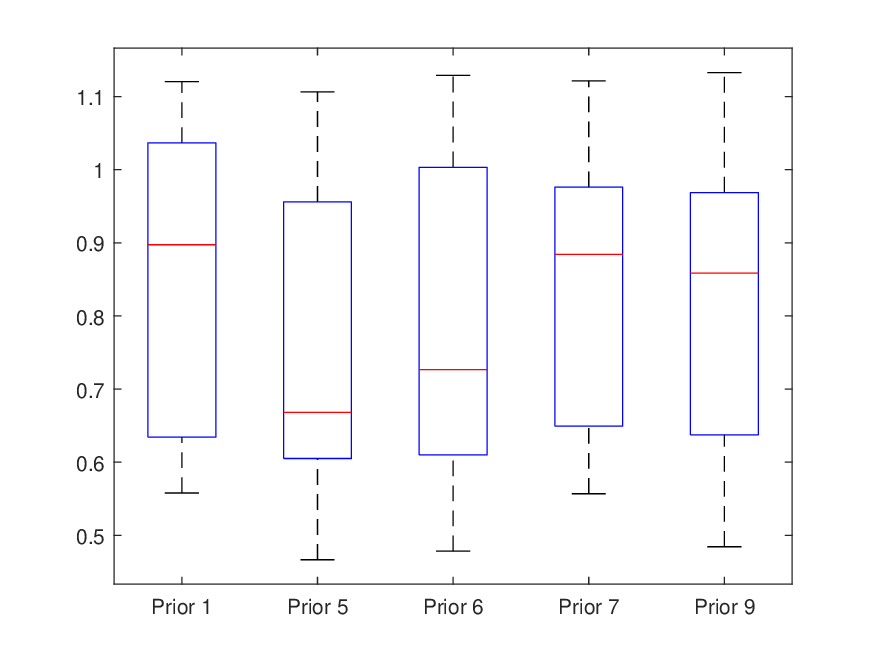}}
\subfigure[Kernel SM, N = 10]{\includegraphics[width=0.49\textwidth]{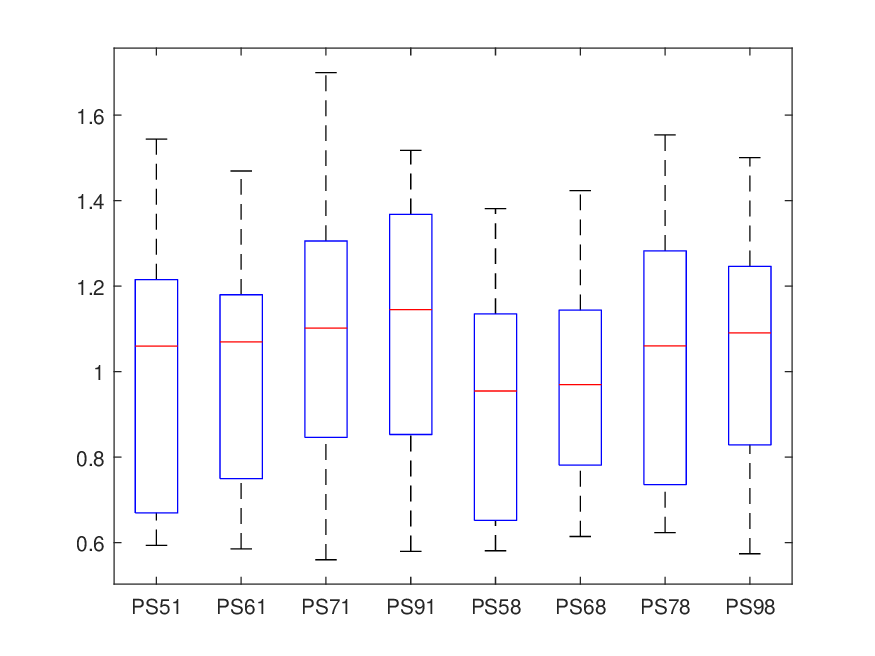}}
\subfigure[Kernel LP, N = 20]{\includegraphics[width=0.49\textwidth]{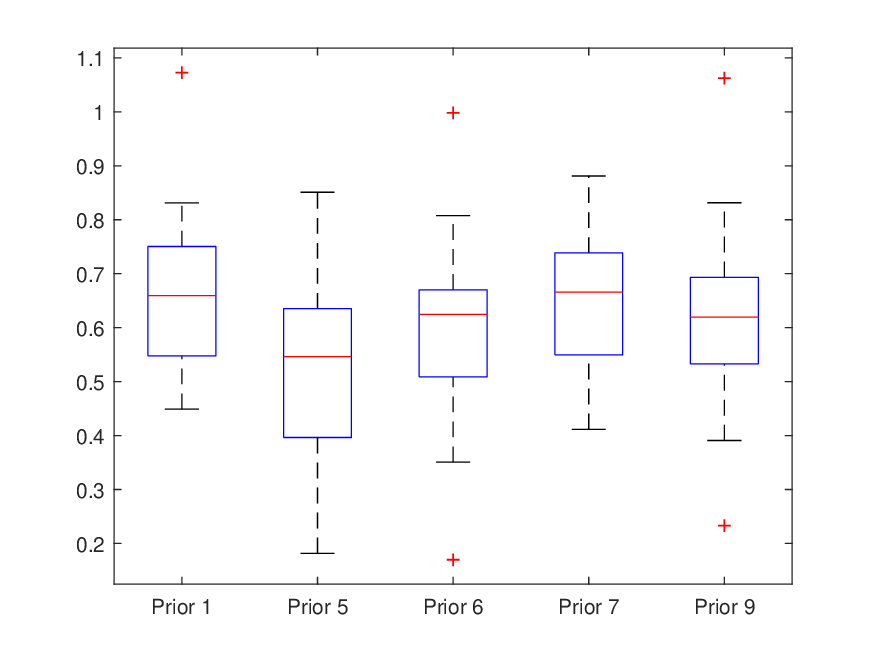}}
\subfigure[Kernel SM, N = 20]{\includegraphics[width=0.49\textwidth]{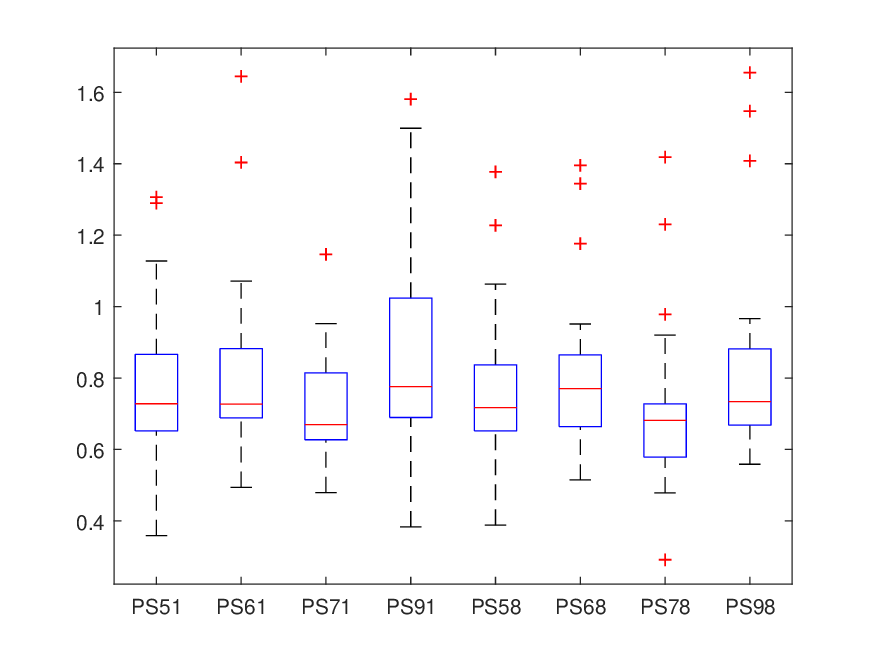}}
\subfigure[Kernel LP, N = 30]{\includegraphics[width=0.49\textwidth]{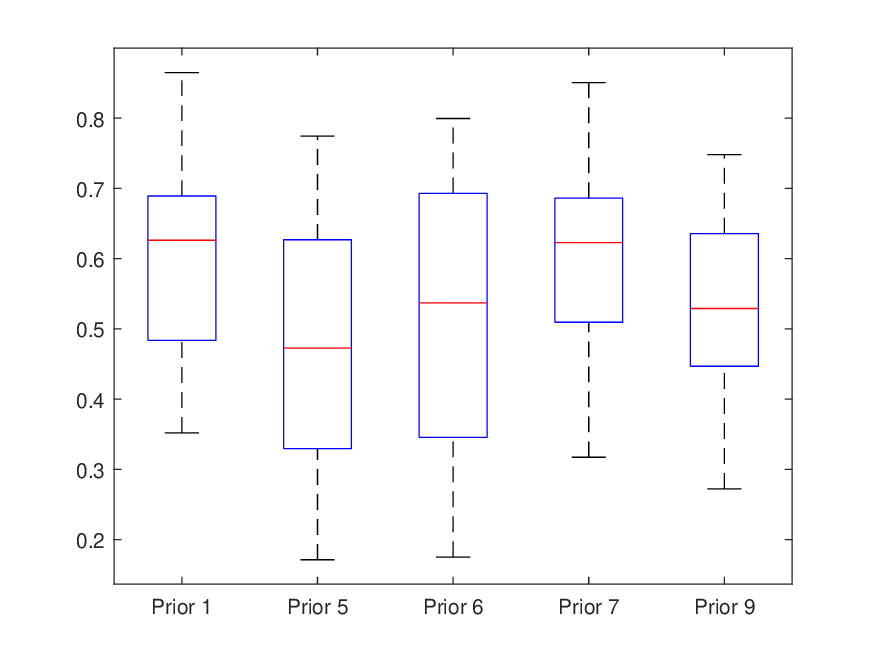}}
\subfigure[Kernel SM, N = 30]{\includegraphics[width=0.49\textwidth]{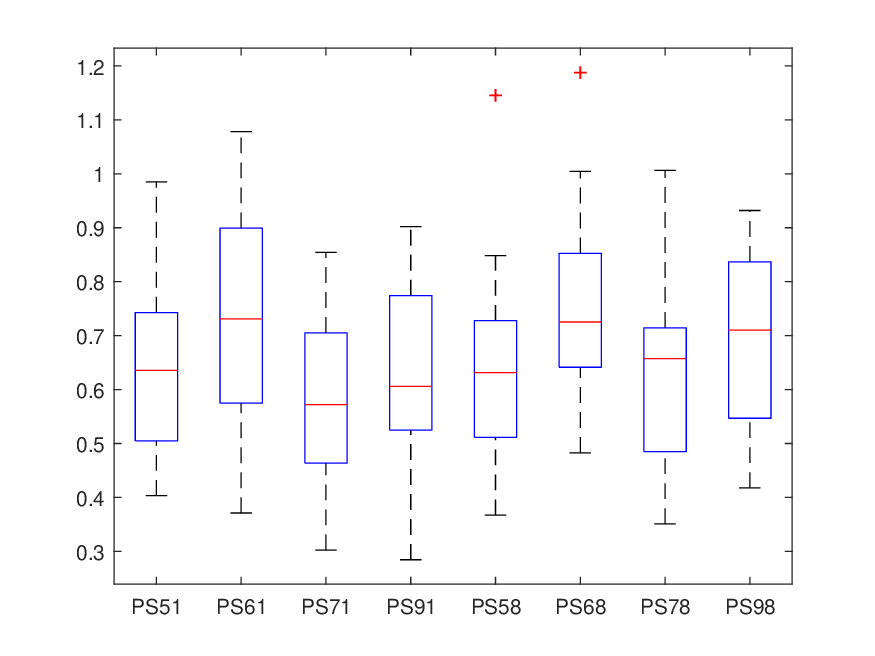}}
\caption{Boxplots of the RMSEs for 20 replications. (a), (c) and (e): kernel LP;
(b), (d) and (f): kernel SM. (a) and (b): N = 10; (c) and (d): N = 20; (e) and (f): N = 30.}
\label{fig:boxplot}
\end{figure}

In order to test whether different priors have significant impact on the prediction accuracy,
the Kruskal-Wallis analysis of variance \cite{meyer2006expanded} is conducted and
the p-values of the test are presented in Table \ref{tab:p-values}. The boxplots of the RMSEs
 for the 20 replications for each case are demonstrated in Figure \ref{fig:boxplot}.
The test shows that, provided that the other settings are kept the same, the priors for the hyperparameter have no significant impact on the prediction accuracy of the GPR model at $5\%$ significant level.

\begin{table}[htbp]
  \centering
  \caption{The p-values of Kruskal-Wallis ANOVA for different N}
    \begin{tabular}{rrrr}
    \toprule
         & N=10  & N=20  & N=30 \\
    \midrule
    LP & 0.722 & 0.055 & 0.164 \\
    SM & 0.583 & 0.278 & 0.077 \\
    \bottomrule
    \end{tabular}%
  \label{tab:p-values}%
\end{table}%

\section{Conclusion}
In this paper, we conducted the empirical studies to investigate the influences of various prior distributions of the initial hyperparameters in GPR models on the parameter estimation and the predictability of the models when numerical optimisation of likelihood function was utilised. Nine commonly used priors and four kernels, including two basic kernels (SE and PER) and two complex kernels (LP and SM), were considered.

The results by the simulated experiments show that the sensitivity of the hyperparameter estimation depends on the choice
of kernels. The estimates for SE kernel are robust regardless of the prior distributions, whilst they are very different using different priors for PER kernel which implies that the prior distributions have huge impact on the estimates of the parameters. However, it is interesting to see that the GPR models always perform well in terms of predictability, despite the poor estimates of the hyperparameters in some cases. Particularly the performances of the GPR models using various priors
are consistently comparable with that of the true time series model in terms of prediction accuracy.
The real data example confirms that the priors for the hyperparameter have no significant impact on the predictability of the GPR model.
Overall, prior distributions of the hyperparameters have little impact on the performance of GPR models, which implies that simple priors, such as the Uniform distribution in an appropriate range, may be sufficient in GPR modelling in terms of predictability.
This study could provide useful guidances to researchers and practitioners using GP as a modelling tool.

It is noted that in terms of evaluating the influences of prior distributions on the performance of GPR models, the study in this paper is far from comprehensive. A wider range of priors and kernels need to be considered, as well as more complex data, including real data. Theoretical analysis may also be of importance because it is not feasible for numerical examples to cover all scenarios.
\section*{Acknowledgement}
The authors are immensely grateful to Prof A. Gorban, University of Leicester, for his helpful comments on the earlier version of the manuscript. The authors also thank all the reviewers for their constructive suggestions and comments.

\section*{References}
\bibliography{mybibfile}

\begin{thebibliography}{10}
\expandafter\ifx\csname url\endcsname\relax
  \def\url#1{\texttt{#1}}\fi
\expandafter\ifx\csname urlprefix\endcsname\relax\def\urlprefix{URL }\fi
\expandafter\ifx\csname href\endcsname\relax
  \def\href#1#2{#2} \def\path#1{#1}\fi

\bibitem{cite:RCE}
C.~E. Rasmussen, Evaluation of {G}aussian processes and other methods for
  non-linear regression, University of Toronto, 1999.

\bibitem{cite:neal}
R.~M. Neal, Bayesian learning for neural networks, Vol. 118, Springer Science
  \& Business Media, 2012.

\bibitem{cite:MacKay}
D.~J. MacKay, Gaussian processes-a replacement for supervised neural networks?

\bibitem{cite:WCKI}
C.~K. Williams, C.~E. Rasmussen, Gaussian processes for regression, in:
  Advances in neural information processing systems, 1996, pp. 514--520.

\bibitem{he2011single}
H.~He, W.-C. Siu, Single image super-resolution using {G}aussian process
  regression, in: Computer Vision and Pattern Recognition (CVPR), 2011 IEEE
  Conference on, IEEE, 2011, pp. 449--456.

\bibitem{ko2009gp}
J.~Ko, D.~Fox, Gp-bayesfilters: Bayesian filtering using {G}aussian process
  prediction and observation models, Autonomous Robots 27~(1) (2009) 75--90.

\bibitem{van2002bayesian}
T.~Van~Gestel, J.~A. Suykens, G.~Lanckriet, A.~Lambrechts, B.~De~Moor,
  J.~Vandewalle, Bayesian framework for least-squares support vector machine
  classifiers, {G}aussian processes, and kernel fisher discriminant analysis,
  Neural computation 14~(5) (2002) 1115--1147.

\bibitem{yu2012online}
J.~Yu, Online quality prediction of nonlinear and non-{G}aussian chemical
  processes with shifting dynamics using finite mixture model based gaussian
  process regression approach, Chemical Engineering Science 82 (2012) 22--30.

\bibitem{shamshirband2015comparative}
S.~Shamshirband, K.~Mohammadi, L.~Yee, D.~Petkovi{\'c}, A.~Mostafaeipour, A
  comparative evaluation for identifying the suitability of extreme learning
  machine to predict horizontal global solar radiation, Renewable and
  Sustainable Energy Reviews 52 (2015) 1031--1042.

\bibitem{jahangirzadeh2014cooperative}
A.~Jahangirzadeh, S.~Shamshirband, S.~Aghabozorgi, S.~Akib, H.~Basser, N.~B.
  Anuar, M.~L.~M. Kiah, A cooperative expert based support vector regression
  ({Co-ESVR}) system to determine collar dimensions around bridge pier,
  Neurocomputing 140 (2014) 172--184.

\bibitem{shamshirband2015soft}
S.~Shamshirband, M.~Goci{\'c}, D.~Petkovi{\'c}, H.~Saboohi, T.~Herawan,
  M.~L.~M. Kiah, S.~Akib, Soft-computing methodologies for precipitation
  estimation: a case study, IEEE Journal of Selected Topics in Applied Earth
  Observations and Remote Sensing 8~(3) (2015) 1353--1358.

\bibitem{damianou2013deep}
A.~Damianou, N.~Lawrence, Deep {G}aussian processes, in: Artificial
  Intelligence and Statistics, 2013, pp. 207--215.

\bibitem{mattos2015recurrent}
C.~L.~C. Mattos, Z.~Dai, A.~Damianou, J.~Forth, G.~A. Barreto, N.~D. Lawrence,
  Recurrent {G}aussian processes, in: International Conference on Learning
  Representations (ICLR), 2015.

\bibitem{wilson2013gaussian}
A.~Wilson, R.~Adams, Gaussian process kernels for pattern discovery and
  extrapolation, in: Proceedings of the 30th International Conference on
  Machine Learning (ICML-13), 2013, pp. 1067--1075.

\bibitem{cite:Rbook}
C.~E. Rasmussen, C.~K. Williams, Gaussian processes for machine learning,
  Vol.~1, MIT press Cambridge, 2006.

\bibitem{cite:monte}
R.~M. Neal, Monte carlo implementation of {G}aussian process models for
  bayesian regression and classification, arXiv preprint physics/9701026.

\bibitem{cite:BBS}
S.~Brahim-Belhouari, A.~Bermak, Gaussian process for nonstationary time series
  prediction, Computational Statistics and Data Analysis 47~(4) (2004)
  705--712.

\bibitem{mackay1998introduction}
D.~J. MacKay, Introduction to {G}aussian processes, NATO ASI Series F Computer
  and Systems Sciences 168 (1998) 133--166.

\bibitem{cite:autokernel}
D.~Duvenaud, J.~R. Lloyd, R.~Grosse, J.~B. Tenenbaum, Z.~Ghahramani, Structure
  discovery in nonparametric regression through compositional kernel search,
  arXiv preprint arXiv:1302.4922.

\bibitem{wilson2014covariance}
A.~G. Wilson, Covariance kernels for fast automatic pattern discovery and
  extrapolation with {G}aussian processes, Ph.D. thesis, University of
  Cambridge (2014).

\bibitem{cite:twins}
S.~Roberts, M.~Osborne, M.~Ebden, S.~Reece, N.~Gibson, S.~Aigrain, Gaussian
  processes for time-series modelling, Phil. Trans. R. Soc. A 371~(1984) (2013)
  20110550.

\bibitem{lambert2005vague}
P.~C. Lambert, A.~J. Sutton, P.~R. Burton, K.~R. Abrams, D.~R. Jones, How vague
  is vague? {A} simulation study of the impact of the use of vague prior
  distributions in {MCMC} using {WinBUGS}, Statistics in Medicine 24~(15)
  (2005) 2401.

\bibitem{gelman2006prior}
A.~Gelman, et~al., Prior distributions for variance parameters in hierarchical
  models (comment on article by browne and draper), Bayesian Analysis 1~(3)
  (2006) 515--534.

\bibitem{rosenkrantz2012jaynes}
R.~D. Rosenkrantz, ET Jaynes: Papers on probability, statistics and statistical
  physics, Vol. 158, Springer Science \& Business Media, 2012.

\bibitem{bammes2012direct}
B.~E. Bammes, R.~H. Rochat, J.~Jakana, D.-H. Chen, W.~Chiu, Direct electron
  detection yields {cryo-EM} reconstructions at resolutions beyond 3/4
  {Nyquist} frequency, Journal of structural biology 177~(3) (2012) 589--601.

\bibitem{wang2015gaussian}
B.~Wang, T.~Chen, Gaussian process regression with multiple response variables,
  Chemometrics and Intelligent Laboratory Systems 142 (2015) 159--165.

\bibitem{tang2009statistical}
Q.~Tang, Y.~Chen, C.~J. Zhou, T.~Chen, Y.~Yang, Statistical modelling and
  analysis of the aerobic oxidation of benzyl alcohol over {K--Mn/C} catalysts,
  Catalysis letters 128~(1-2) (2009) 210--220.

\bibitem{meyer2006expanded}
J.~P. Meyer, M.~A. Seaman, Expanded tables of critical values for the
  {Kruskal-Wallis H} statistic, in: annual meeting of the American Educational
  Research Association, San Francisco, 2006.

\end{thebibliography}
\end{document}